\documentclass[conference]{IEEEtran}
\IEEEoverridecommandlockouts
\usepackage{cite}
\usepackage{amsmath,amssymb,amsfonts}
\usepackage{algorithmic}
\usepackage{graphicx}
\usepackage{textcomp}
\usepackage{xcolor}
\usepackage{booktabs}

\def\BibTeX{{\rm B\kern-.05em{\sc i\kern-.025em b}\kern-.08em
    T\kern-.1667em\lower.7ex\hbox{E}\kern-.125emX}}
\begin{document}

\title{Incorporating Human-Inspired Ankle Characteristics in a Forced-Oscillation-Based Reduced-Order Model for Walking}

\author{\IEEEauthorblockN{Chathura Semasnighe}
\IEEEauthorblockA{\textit{Department of Mechanical Engineering} \\
\textit{University of Denver}\\
Denver, CO, USA\\
Chathura.Semasinghe@du.edu}
\and
\IEEEauthorblockN{Siavash Rezazadeh}
\IEEEauthorblockA{\textit{Department of Mechanical Engineering} \\
\textit{University of Denver}\\
Denver, CO, USA\\
Siavash.Rezazadeh@du.edu}
}

\maketitle

\begin{abstract}
This paper extends the forced-oscillation-based reduced-order model of walking to a model with ankles and feet. A human-inspired paradigm was designed for the ankle dynamics, which results in improved gait characteristics compared to the point-foot model. In addition, it was shown that while the proposed model can stabilize against large errors in initial conditions through combination of foot placement and ankle strategies, the model is able to stabilize against small perturbations without relying on the foot placement control and solely through the designed proprioceptive ankle scheme. This novel property, which is also observed in humans, can help in better understanding of anthropomorphic walking and its stabilization mechanisms.
\end{abstract}

\section{Introduction}

%Bipedal robots have gained increasing attention in recent years for human-oriented applications due to their anthropomorphic structural design and their ability to traverse diverse terrains \cite{Tameemi_Review, Liao_V_DSLIP}.
Reduced-order models of legged systems aim to capture the fundamental traits of locomotion within a lower-dimensional framework. Locomotion control, gait generation, and stability remain key challenges in both biomechanical and robotic locomotion, and the complexity and nonlinear dynamics of these systems have motivated the use of simplified models. One of the oldest and most essential models of bipedal walking is the Inverted Pendulum Model (IPM), which has long been used for control and stability analyses of bipedal locomotion \cite{Hemami_IPmodel}. Building on this and by incorporating the Zero Moment Point (ZMP) concept proposed by Vukobratović and Juričić \cite{Vukobratovic_ZMP}, Kajita et al. \cite{Kajita_PatternGen, Kajita_GaitControl} extended the IPM to the Linear Inverted Pendulum Model (LIPM), a simplified model widely used for humanoid gait generation and control. The LIPM is among the most prevalent templates for bipedal walking, as it enables real-time control through fast, convex trajectory generation of the point mass, which is also the center of mass (CoM) \cite{Kajita_GaitControl, Liu_LIPMcontrol}. However, the assumption of constant leg length constrains vertical CoM motion, allowing only horizontal dynamics. With this restriction, the vertical ground reaction force (GRF) remains effectively constant and equal to the weight, which prevents the reproduction of the human-like M-shaped vertical GRF profiles \cite{Englsberger_DCM, Liu_Wensing_3DdualSLIP}. These limitations raise concerns about the ability of the LIPM template to capture essential characteristics of anthropomorphic walking. Although the original IPM produces attributes more closely resembling human walking, notable differences remain, particularly in GRFs and CoM displacement \cite{Geyer_SLIPwalkRun}. More recently, by adding a force actuator to the IPM and applying trajectory optimization methods, these discrepancies have been addressed \cite{rebula2015cost}; however, this extended model is primarily valuable for understanding biomechanical costs, while its application to robotic systems remains an open question.

Another widely adopted reduced-order model is the Spring-Loaded Inverted Pendulum (SLIP) model, which represents the system as a single point mass supported by one or two massless elastic legs. Originally developed for running, it was later extended to capture key features of human walking \cite{BLICKHAN_SLIP, Geyer_SLIPwalkRun, Meng_Yu_SLIPwalk}. Unlike IPM and LIPM, the energy storage and release mechanism provided by its compliance enables this model to reproduce human-like gait characteristics, including the double-hump (M-shaped) GRF profile and realistic CoM trajectories \cite{Geyer_SLIPwalkRun, Zaytsev_SLIPmodel}.

Over the years, several variations of the SLIP model have been proposed to improve walking performance and provide closer approximations of human locomotion characteristics. The Variable SLIP (V-SLIP) model \cite{Roozing_V-SLIP, Visser_V-SLIP} incorporates active modulation of leg stiffness to achieve a cost of transport comparable to that of human walking, thereby enabling more energy-efficient and robust bipedal locomotion. To account for upper-body dynamics, Sharbafi and Seyfarth introduced the Bipedal Trunk-SLIP (BTSLIP) model \cite{Sharbafi_BTSLIP, Sharbafi_Seyfarth_BTSLIP}, which redirects GRFs to stabilize trunk posture while preserving compliant leg mechanics, resulting in more realistic GRF profiles and CoM dynamics. Pelit et al.\cite{Pelit_SLIP-SL} extended the framework to the SLIP with Swing Legs (SLIP-SL) model, which incorporates passive swing-leg dynamics without requiring separate swing-leg trajectory planning. This extension enhances stability and cyclic locomotion by producing human-like gait timing and step lengths. Geyer et al. \cite{Geyer_SLIPwalkRun} proposed a simplified Bipedal SLIP (BSLIP) model capable of generating stable walking using different combinations of overall leg stiffness and attack angles. Hao et al. \cite{Hao_D-SLIP} further advanced the BSLIP framework by introducing a damper in parallel with the spring on each massless leg, resulting in the D-SLIP model, which allows evaluation of the effects of hip actuator torque on the energetics and stability of bipedal walking.

\begin{figure}
	\centering\includegraphics[width=0.45\linewidth]{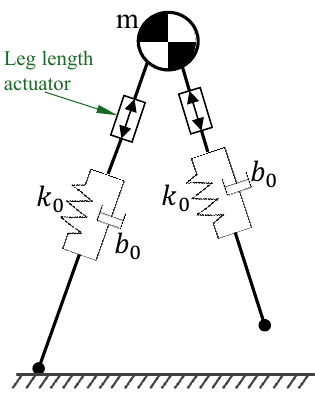}
	\caption{The forced oscillation-based reduced-order model; an extension of SLIP with damping and leg length actuation with enhanced stability properties \cite{Rezazadeh_Hurst_2020}. \label{fig_1_Models}}
\end{figure}

However, the SLIP model and its variations, while useful in capturing human locomotion traits, are highly simplified and underactuated. Hence their deployment to robots remains challenging, often because of underactuation and merely marginal stability in the template level \cite{rezazadeh2015toward, Vejdani_Geyer_Hurst2025}. Notwithstanding, the forced-oscillation-based variation of SLIP (Fig. \ref{fig_1_Models}) is a template that has resulted in highly stable gaits on bipedal robots \cite{Rezazadeh_Hurst_2015, Rezazadeh_Hurst_2020}. In this model, a damper and a leg length actuator for setting the spring's neutral length are added to the bipedal SLIP. The stability of this model and its wide basin of attraction have been proved and its human-like characteristics have been presented in \cite{Rezazadeh_Hurst_2020}. As such, adopted this paradigm as the foundation for an extension aimed at incorporating more anthropomorphic traits.

%%%%%%% SINGLE COLUMN FIGURE

A significant shortcoming in the aforementioned point-foot models is that they neglect the kinematics and dynamics of the foot and ankle, which play essential roles in human walking. In particular, ankle flexion contributes a substantial amount of positive mechanical work for forward progression and provides vertical support to sustain body weight \cite{Ankle_MechImpedance, AnklePlantarFl_model}. Therefore, incorporating an ankle joint and a finite-sized foot into an extended reduced-order model can potentially enhance stability and robustness in bipedal walking. In addition, the ankle push-off in such models can produce more realistic GRF profiles, CoM dynamics closer to that of humans, human-like forward center-of-pressure (CoP) travel, and ankle behavior, which are features that point-foot templates fail to capture \cite{Xie_AnkleModel_21, Whittington_RollerFeetModel}. An example of such an extension is proposed in \cite{Xie_AnkleModel_21}, where the V-SLIP model was equipped with a 1-DoF ankle joint and a finite-sized foot (VSLIP-FF) for each leg, enabling compliant bipedal walking in complex environments. Their approach employed a finite state machine–based gait planner combined with a differential evolution (DE) optimizer to generate periodic gaits. The gait planning method incorporated an adaptive leg stretching and contracting strategy inspired by human walking, with pre-determined footholds used to update step length for placing the foot on known targets. This strategy was controlled by a PD controller and its gains were selected through offline tuning/optimization to realize a periodic gait. Although this method can generate stable gaits, its reliance on offline tuning and optimization limits its flexibility. This challenges the online adaptability and robustness of bipedal walking, particularly for achieving anthropomorphic stance mechanics, real-time foot placement, and step regulation in response to speed variations or disturbances. More importantly, the stability in this framework does not inlcude the distinct separation between ankle and foot placement strategies \cite{vlutters2016center}. In general, many SLIP model variations ultimately produce gaits dictated by the natural dynamics of the system, without explicitly comparing to anthropomorphic walking gaits \cite{Vejdani_Geyer_Hurst2025}.

In this paper, we propose an extended version of the forced-oscillation-based reduced-order model for human-like walking, where an ankle and a foot are added. A human-inspired scheme is designed for the ankle characteristics and all the tunable parameters are selected using human biomechanical information. To evaluate the performance of the proposed model, we compare its results against both human data and the point-foot forced-oscillation-based model and show how the ankle stabilization can work without foot placement regulation, similar to humans.

The rest of this paper is organized as follows. Section II describes the proposed model and its designed characteristics. Section III details the model parameters and the simulation process. Section IV presents the simulation results including comparisons with the point-foot model and the human data. These results and comparisons are discussed in Section V. Finally, Section VI concludes the paper and outlines directions for future works.

%%%%%%%%%%%%%%%%%%%%%%%%%%%%%%%%%%%%%%%%%%%%%%%%%%%%%%%%%%%%%%%%%%%%%%
\section{The Proposed Reduced-Order Model and Its Analysis}

\subsection{The Model}
\begin{figure}
	\centering\includegraphics[width=0.9\linewidth]{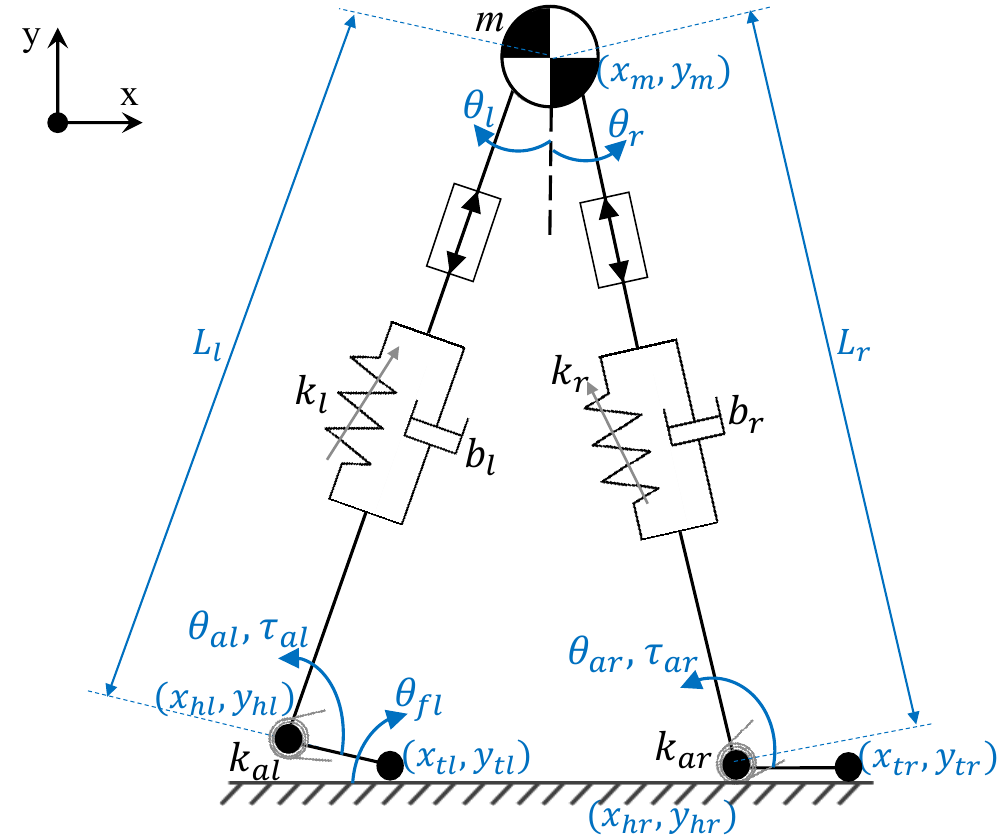}
	\caption{Detailed description of the proposed model and the associated variables.\label{fig_2_DVSLIP_FF}}
\end{figure}

The proposed reduced-order model consists of a point mass, representing the total body mass $m$, and two compliant massless legs with spring stiffness and damping coefficients that can be variable. Each leg is connected to a finite-sized massless foot of length of $L_f$ through a 1-DoF ankle joint, as illustrated in Fig. \ref{fig_2_DVSLIP_FF}. The legs are indexed by $i \in \{l, r\}$ for left and right, respectively. With this definition, $L_i$ will denote the actual leg length, $k_i$ the leg stiffness, $b_i$ the damping coefficient of the leg, and $k_{ai}$ the stiffness of the ankle joint for the leg $i$.

%%%%%%% SINGLE COLUMN FIGURE

To represent the hybrid dynamics of walking, we employ a finite state machine with a set of discrete modes (phases) which includes single-support, double-support, and push-off, along with their associated transition rules. Each phase is governed by its own continuous dynamics, and transitions between modes are triggered by predefined events. The details of these transitions will be provided in the next subsection.

The considered state vector includes the position of the point mass $P_m = [x_m, y_m]$, its velocity $V_m = [\dot{x}_m, \dot{y}_m]$, position of the left heel $P_{hl} = [x_{hl}, y_{hl}]$, and right heel $P_{hr} = [x_{hr}, y_{hr}]$. For continuous dynamics, control strategy implementation, and gait analysis, dependent parameters are derived for each leg within the active mode using the state vector and independent variables. These include the leg angle relative to the vertical axis $\theta_i$, the ankle angle between the leg and foot $\theta_{ai}$, the ankle torque $\tau_{ai}$, and the toe positions of the left foot $P_{tl} = [x_{tl}, y_{tl}]$ and right foot $P_{tr} = [x_{tr}, y_{tr}]$. All notations are presented in Fig. \ref{fig_2_DVSLIP_FF}.

\subsection{Controlled Aspects of the Model}
While the SLIP model effectively captures many essential traits of walking, it overlooks important aspects such as leg retraction and ankle push-off. Moreover, in the classical SLIP model, the only control variable is the foot placement \cite{ernst2012extension, Vejdani_Geyer_Hurst2025}. As such, we design additional simple schemes to encapsulate these key aspects and enhance the performance of the model.

%Considering all discrete modes, each leg of the model can be classified into two primary control modes: stance and swing. In stance-foot control, two motions are particularly critical: axial motion of the leg (stretching and contracting) and foot rotation around the ankle joint. The leg’s axial motion governs vertical loading and point-mass dynamics, thereby enabling compliant and anthropomorphic stance mechanics. This is also essential for reproducing the double-hump vertical GRF profile found in human walking. Meanwhile, the ankle torque and heel-to-toe rollover of the finite foot, provide push-off that reduces step-to-step redirection losses and supports steady forward progression of the CoM, as emphasized in \cite{Donelan_footPushoff}. In swing-foot control, since there is no leg mass, the most critical factor is foot placement, which plays a central role in maintaining steady gait velocity, ensuring stability, and rejecting external disturbances. Therefore, within the finite state machine framework used to simulate the DVSLIP-FF model, three variable need to be effectively controlled: leg axial dynamics, foot rollover with ankle torque, and swing-foot placement.

\subsubsection{Forced-Oscillation Scheme for the Axial Motion of the Legs}
\begin{figure}
	\centering\includegraphics[width=1\columnwidth, trim={15mm 0mm 35mm 0}]{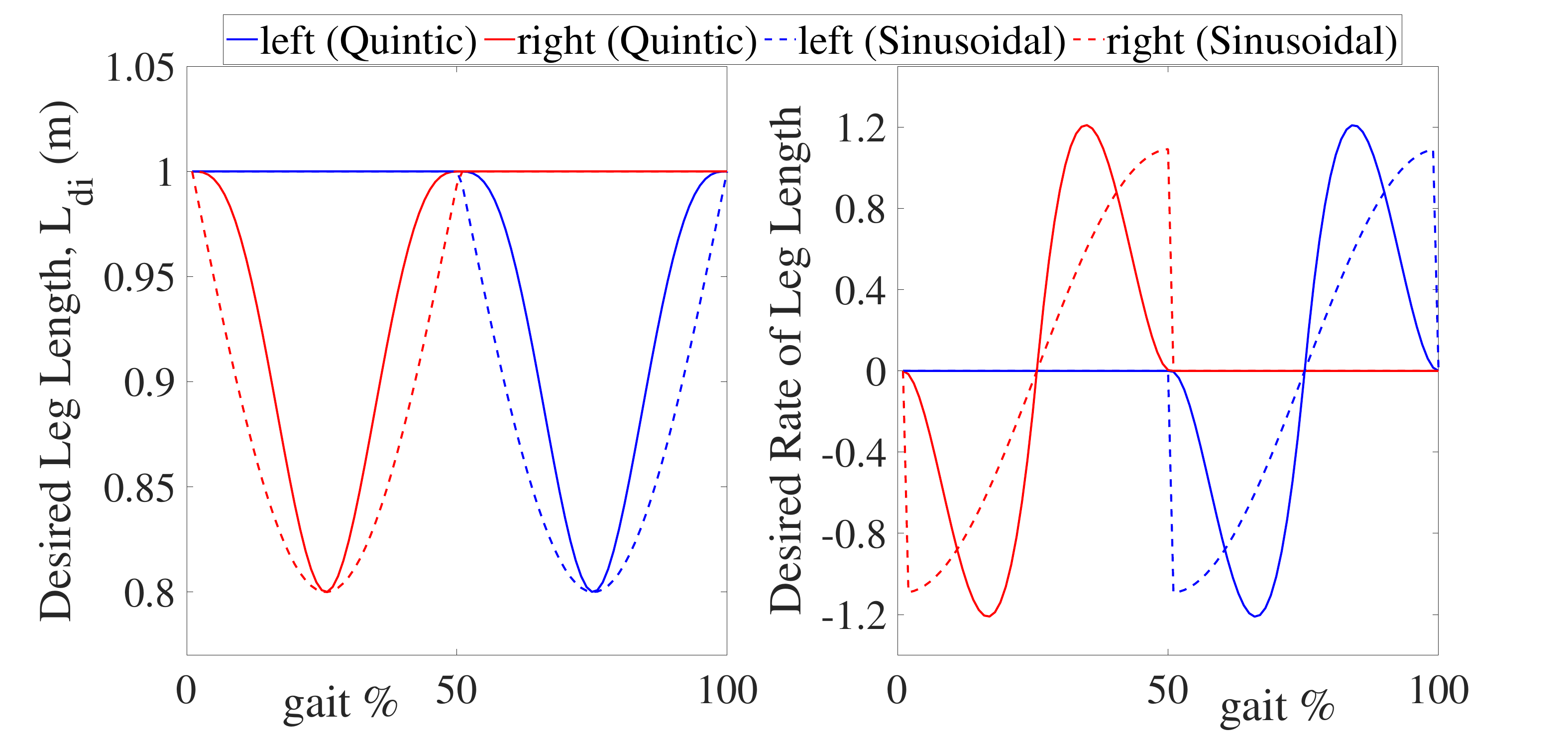}
	\caption{Desired length, $L_{di}$ and its rate of change $\dot{L}_{di}$ for each leg during one gait cycle. Solid line presents output results from function of two piecewise quintic polynomials stitched together with a continuous jerk. The dashed line presents output results from the original  sinusoidal trajectory of \cite{Rezazadeh_Hurst_2015}. %Considered parameters are neutral leg length $L_0 = 1$ m, maximum retraction, $a = 0.2$ m, and stride time or total gait period, $T = 1.15$ s.
		\label{fig_3_desLegTraj}}
	\vspace{4mm}
\end{figure}

In the classical SLIP model, the swing leg motion and especially its flexion-extension is neglected. However, as we showed in \cite{Rezazadeh_Hurst_2015, Rezazadeh_Hurst_2020}, the flexion-extension motion can be designed such that it not only encapsulates this aspect similar to biological locomotion, but also provides a strong stabilizing effect through resulting in forced-oscillations \cite{Rezazadeh_Hurst_2015, Rezazadeh_Hurst_2020}. The effectiveness of this scheme was theoretically proven and experimentally validated with the ATRIAS robot, showing robustness against various perturbations \cite{Rezazadeh_Hurst_2015, Rezazadeh_Hurst_2020}. The neutral spring length in this scheme is defined by a half-sinusoidal function in which the length is held constant during the first half of the stride (approximately, the stance phase), while in the second half the leg retracts and subsequently returns to its maximum length following the sinusoidal shape. To resolve the discontinuities that can arise from this scheme in the velocity level, we used two piecewise quintic polynomials stitched together with continuous jerk (i.e., $C^3$ continuity) to generate both the desired leg length, $L_{di}$ and its rate of change, $\dot{L}_{di}$ throughout the stride. These trajectories are then used to calculate the axial leg force (spring-damper force) for the model dynamics. Fig. \ref{fig_3_desLegTraj} illustrates the leg-length trajectories for each leg during a full gait cycle using our proposed control scheme for leg's axial motion. Note that the leg forces in the swing leg will be zero due to both leg and foot being massless. Thus, the stance leg force $F_i$ can be obtained as:
\begin{equation}\label{eqn:1}
	%\mathbf{q} = -k\nabla T
	F_i = k_i(L_{di} - L_i) + b_i(\dot{L}_{di} - \dot{L}_i), \hspace{8mm}  i\in\{l, r\},
\end{equation}
\noindent where $k_i$ is the stiffness of the leg spring, $b_i$ is the damping coefficient, $L_i$ and $\dot{L}_i$ are the actual leg length and its rate of change, respectively.

%%%%%%% SINGLE COLUMN FIGURE

%%%%%%% SINGLE COLUMN FIGURE
\begin{figure}
	\centering\includegraphics[width=0.65\linewidth]{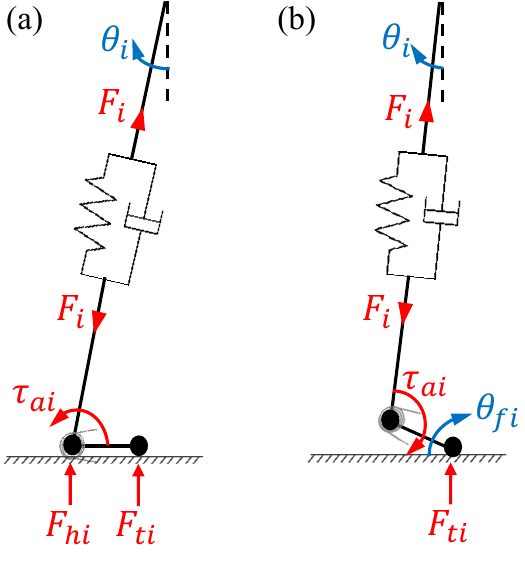}
	\caption{Leg Dynamics, (a) For stance foot in contact with ground through both heel and toe (b) For stance foot at it's push-off \label{fig_4_footDyn}}
\end{figure}

\subsubsection{Ankle Dynamics}
We neglect the small plantarflexion phase after touchdown and assume the heel and toe touch the ground simultaneously at the start of the stance phase. During stance, we consider two different paradigms for the ankle based on whether the heel is on the ground or not. When both heel and toe are on the ground, the ankle acts as a spring resisting the dorsiflexion motion. This is inspired by the observations from human walking during this phase \cite{Shamaei_ankleStiffness}. Considering the free-body diagram of the stance leg with both heel and toe in contact with the ground as shown in Fig. \ref{fig_4_footDyn}(a), the reaction forces on the heel $F_{hi}$ and the toe $F_{ti}$ can be derived as follows:
\begin{equation} \label{eqn:2}
	\begin{split}
		& F_{hi} = F_i\cos\theta_i - \tau_{ai}(\frac{\sin\theta_i}{L_i} + \frac{1}{L_f})\\
		& F_{ti} = \frac{\tau_{ai}}{L_f}
	\end{split}
\end{equation}
%\noindent where the ankle joint moment $\tau_{ai}$ is calculated by $k_{ai}(\pi/2 - \theta_{ai})$. The trigger or starting point of push-off initiates when the $F_{hi}$ passes zero from the positive value. 

\begin{figure}
	\centering\includegraphics[width=0.75\linewidth]{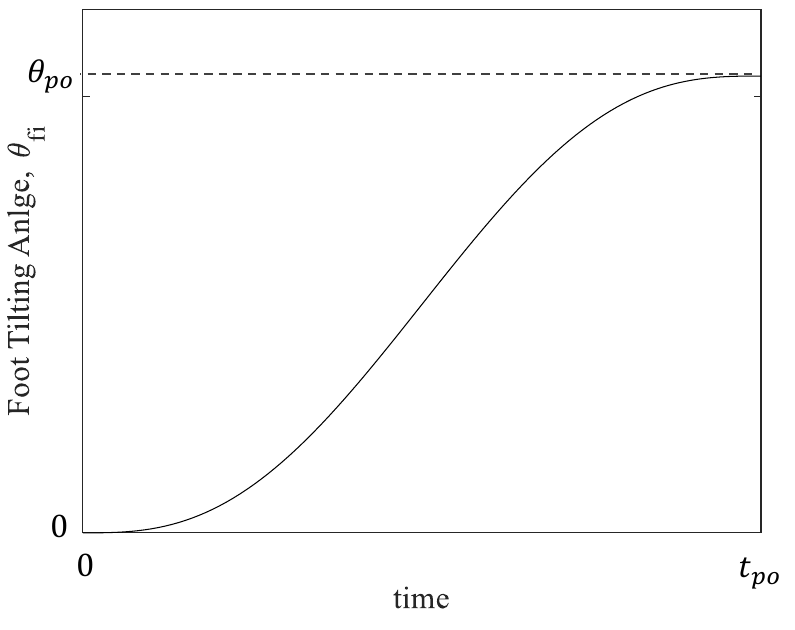}
	\caption{An example of bio-inspired feedforward trajectory of foot tilting at push-off. \label{fig_5_footTraj}}
\end{figure}

Once $F_{hi}$ crosses zero, the heel comes off the ground and the second phase (push-off) is started. For this phase, instead of simulating the ankle as a spring with neutral angle greatly changing compared with the dorsiflexion phase (as in \cite{Shamaei_ankleStiffness}), a bioinspired feedforward foot angle trajectory is used, which, as we will see, will help in proprioceptive stabilization. An example of such a foot angle trajectory $\theta_{fi}$ is shown in Fig. \ref{fig_5_footTraj}, where $\theta_{po}$ is the maximum foot tilting angle and $t_{po}$ is the considered push-off duration. 

%Based on the literature found on foot tilting angle from the ground at ankle push-off during walking for both humans and assitive robotic devices \cite{Vette_footAngle_Humans, Yeung_footAngle_Orthosis}, also based on human data analysis which discussed in the next chapters from publicly available walking dataset from \cite{Gregg_dataset}, a feedfoward trajectory of the foot tilting angle is considered for achieving bio-inspired ankle push-off motion. 

When the heel is off the ground and the toe is in contact, the required ankle moment for a given foot angle can be determined from:
%For allowing foot to follow this feedforward trajectory, using the free-body diagram shown in fig. \ref{fig_4_footDyn}(b), the ankle joint torque, $\tau_{ai}$ can be formulated for the leg dynamics at push-off stage as follows:
\begin{equation} \label{eqn:3}
	\tau_{ai} = \frac{F_iL_f\cos(\theta_{fi}+\theta_i)}{1+ \frac{L_f}{L_i}\sin(\theta_{fi}+\theta_i)} %\hspace{8mm} i\in[l, r]
\end{equation}
\noindent The toe force then can be calculated as:
\begin{equation} \label{eqn:4}
	F_{ti} = F_i\cos\theta_i - \frac{\tau_{ai}\sin\theta_i}{L_i}%, \hspace{8mm}  i\in[l, r]
\end{equation}
\noindent The push-off phase ends when $F_{ti}$ crosses zero, which will be equivalent to the start of the swing phase.

%%%%%%% SINGLE COLUMN FIGURE

%%%%%%%%%%%%%%%%%  begin two column figure  %%%%%%%%%%%%%%%%%%%%%%%%%%%%%%%%%%%%
\begin{figure*}
	\centering\includegraphics[width=0.9\linewidth]{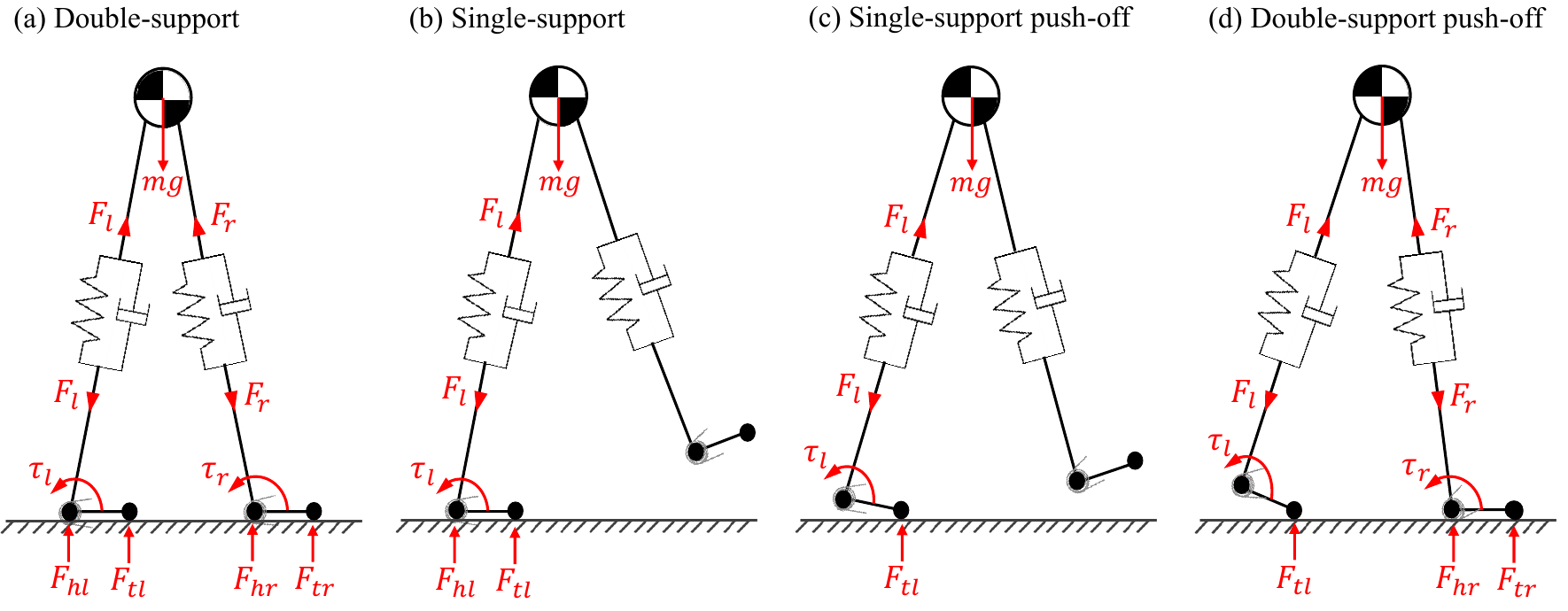}
	\caption{Four phases of our model during a walking cycle: (a) Double-support phase – both feet in ground contact through heel and toe; (b) Single-support phase – one foot in ground contact through heel and toe while the other swings; (c) Single-support push-off – one foot in ground contact only through the heel while the other swings; (d) Double-support push-off – one foot in full ground contact through heel and toe while the other contacts the ground only through the toe.\label{fig_6_gaitPhases}}
\end{figure*}

\subsubsection{Foot Placement}

As proven in \cite{Rezazadeh_Hurst_2020}, there are a class of foot placement functions based on the mass velocity that can stabilize the point-foot model with the forced-oscillation scheme. The velocity-based foot placement has also been observed in human walking \cite{wang2014stepping, afschrift2021similar}. Hence, these factors motivated us to use this scheme for our new reduced-order model as well.

Instead of assuming instantaneous foot swing (as in SLIP), we implement a time-parameterized trajectory generated by a quintic spline profile. This extension enables the model to account for kinematic and actuator constraints, such as limited leg angular velocities and accelerations, which are inherent in real-world systems. Furthermore, the quintic spline formulation ensures that the swing foot achieves touchdown with a small relative velocity, thereby enhancing placement robustness and overall gait stability. As such, these extension not only improve physical realism but also provide flexibility to adapt to specific joint actuator characteristics when required.

\subsection{Walking Phases and Transitions}

\begin{figure}
	\centering\includegraphics[width=0.9\linewidth]{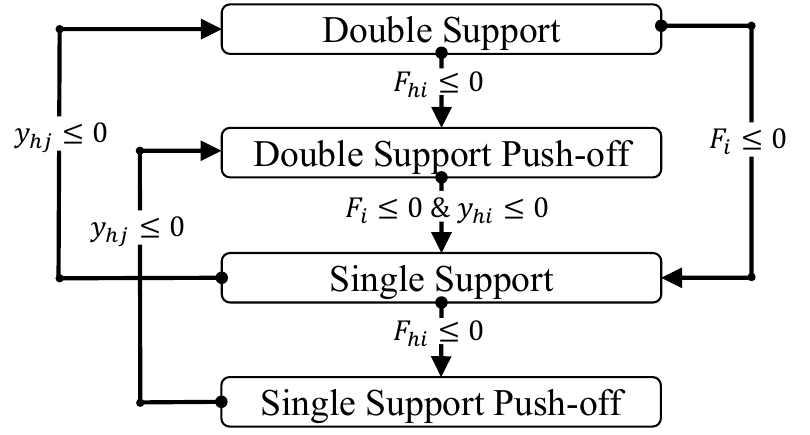}
	\caption{Phase transitions and the corresponding event triggers.\label{fig_8_disModeTrans}}
\end{figure}

In general, bipedal walking has two phases: single-support and double-support. However, incorporating ankle push-off introduces sub-phases that have different dynamics from the normal stance and swing. As such, we consider two separate phases, namely single-support push-off and double-support push-off. The four phases defined in our model are illustrated in Fig. \ref{fig_6_gaitPhases}. Noting that a contact point (i.e., heel or toe) lifts off when its ground reaction force crosses zero, and touches down when its height crosses zero, several possible transitions between the aforementioned phases can occur. These phase transitions are summarized in Fig. \ref{fig_8_disModeTrans}. The heel and toe forces are determined from Eqns. \ref{eqn:2} and \ref{eqn:4}.

%%%%%%% SINGLE COLUMN FIGURE

\section{Simulations}
\subsection{Model Parameters}
%Based on the chosen parameter values, the proposed model may either converge to steady-state locomotion or stumble and fall. Therefore, parameter selection must be performed carefully to avoid overfitting, ensuring that the model does not function reliably only under narrowly tuned conditions. A review of the literature identifies three major approaches for selecting parameters in reduced-order models of bipedal walking.

Parameter selection for simulations must be performed carefully to avoid overfitting and ensure that the model does not function reliably only under narrowly tuned conditions. A review of the literature identifies three major approaches for selecting parameters in reduced-order models of bipedal walking. Approach 1 determines parameters through optimization, by fitting one or more model outputs (such as GRF profiles, CoM trajectories, or energy/cost of transport) to human walking data \cite{Li_humanOpti, Mauersberger_humanOpti}. Although this approach can achieve close agreement with experimental datasets, the resulting parameters often show optimistic bias and perform well only under the given state-space environment and speed. Approach 2 embeds the model within a gait optimization framework to generate dynamically consistent gaits that trade-offs between energy efficiency, stability, and human-like characteristics \cite{Xie_AnkleModel_21, Liao_V_DSLIP}. Despite its effectiveness, this method is also constrained by its limited ability to generalize. As such, in this paper, we do not rely on optimization algorithms for parameter selection, as our primary aim is to propose a new reduced-order model and evaluate its robustness in reproducing human-like walking characteristics. Nevertheless, our model is compatible with these approaches and could be integrated into such frameworks in future works. Approach 3, which we adopt, is to select parameter values from prior literature on SLIP-family models that reproduce fundamental walking mechanics, including realistic vertical GRF profiles and ankle push-off dynamics. This approach enables meaningful comparison with both human datasets and other reduced-order templates, while demonstrating robustness across reasonable parameter ranges.

Table \ref{tab:1-ModParam} summarizes the model parameters selected for generating walking at 1.2 m/s. These values are drawn from \cite{Wensing_variStiffness, Shamaei_ankleStiffness, Geyer_SLIPwalkRun, Martinez_CompliantLeg, Seyfarth_SLIPwalk, Li_humanOpti}, and are further supported by processed human gait data from the publicly available dataset in \cite{Gregg_datasetPaper}, which reports lower-limb kinematics and kinetics from 10 subjects, including walking trials at the same speed.

%%%%%%%%%%%%%%%%%%%  begin  small table  %%%%%%%%%%%%%%%%%%%%%%%%%%%%%%

\begin{table}[h]
	\caption{Model parameters used in simulation trial for achieving walking gait speed of 1.2 m/s\label{tab:1-ModParam}}
	\centering{%
		\begin{tabular}{l c c c}
			\midrule
			Body weight, $m$ (Kg) & \multicolumn{3}{c}{75} \\
			Rest leg length, $L_0$ (m) & \multicolumn{3}{c}{1.00} \\ 
			Stride time, $T$ (s) & \multicolumn{3}{c}{1.15} \\
			Foot angle at push-off, $\theta_{po}$ (deg) & \multicolumn{3}{c}{30} \\ 
			Foot push-off duration, $t_{po}$ (s) & \multicolumn{3}{c}{20\% of Stride time} \\
			Case & 1 & 2 & 3\\
			Leg stiffness, $k_i$ (N/m) &  &  & \\
			\hfill at touchdown ($TD$) & 14000 & 9000 & 14000\\
			\hfill during stance ($SS$) & 14000 & 14000 & 10000\\
			\hfill at push-off ($PO$) & 14000 & 14000 & 14000\\
			Leg damping ratio, $\zeta$ &  &  & \\
			\hfill at touchdown ($TD$) & 0.01 & 0.01 & 0.01\\
			\hfill during stance ($SS$) & 0.08 & 0.08 & 0.08\\
			\hfill at push-off ($PO$) & 0.1 & 0.1 & 0.1\\
			Ankle joint stiffness, $k_{ai}$ (Nm/rad) &  &  & \\
			\hfill at touchdown ($TD$) & 200 & 200 & 200\\
			\hfill during stance ($SS$) & 400 & 440 & 420\\
			\midrule
		\end{tabular}
	}%
\end{table}

%%%%%%%%%%%%%%%%%%%%  end small table %%%%%%%%%%%%%%%%%%%%%%%

\subsection{Model Simulation} 
The stiffness of the compliant leg and ankle joint significantly influence the performance of the proposed model. In the prior literature, several studies have investigated stiffness profiles in simplified models to reproduce human-like walking characteristics. Geyer et al. \cite{Geyer_SLIPwalkRun} demonstrated the importance of compliant legs for capturing fundamental features of human walking, using constant leg stiffness values of 14 and 20 kN/m. In contrast, Kelly and Wensing \cite{Wensing_variStiffness} emphasized the necessity of continuously time-varying leg stiffness within a stride to achieve more accurate human-like GRF matching and CoM tracking, both qualitatively and quantitatively, while also preserving realistic gait event timing. Two main approaches for time-varying stiffness have been reported for compliant legs. The first involves assigning low stiffness at touchdown and gradually increasing it toward mid-stance, which enhances compliance for impact absorption and improves stability on uneven terrain. Zhao et al. \cite{Zhao_varyStiff} supported this approach, showing that vertical knee stiffness fluctuates around zero during the early stance phase. The second approach, adopted by Kelly and Wensing \cite{Wensing_variStiffness}, suggests that human-like bipedal walking tends to exhibit higher stiffness (\textasciitilde 15 kN/m) at touchdown, which subsequently decreases during mid-stance (\textasciitilde 9–14 kN/m). Based on these findings, we evaluate the proposed ankle-augmented model under three cases:

\begin{itemize}
	\item Case 1: constant leg stiffness.
	\item Case 2: variable leg stiffness, starting low at touchdown and gradually increasing to a higher value in mid-stance.
	\item Case 3: variable leg stiffness, starting high at touchdown, decreasing during early stance, and then gradually increasing again toward mid-stance.
\end{itemize}

\begin{figure*}
	\centering\includegraphics[width=0.9\linewidth]{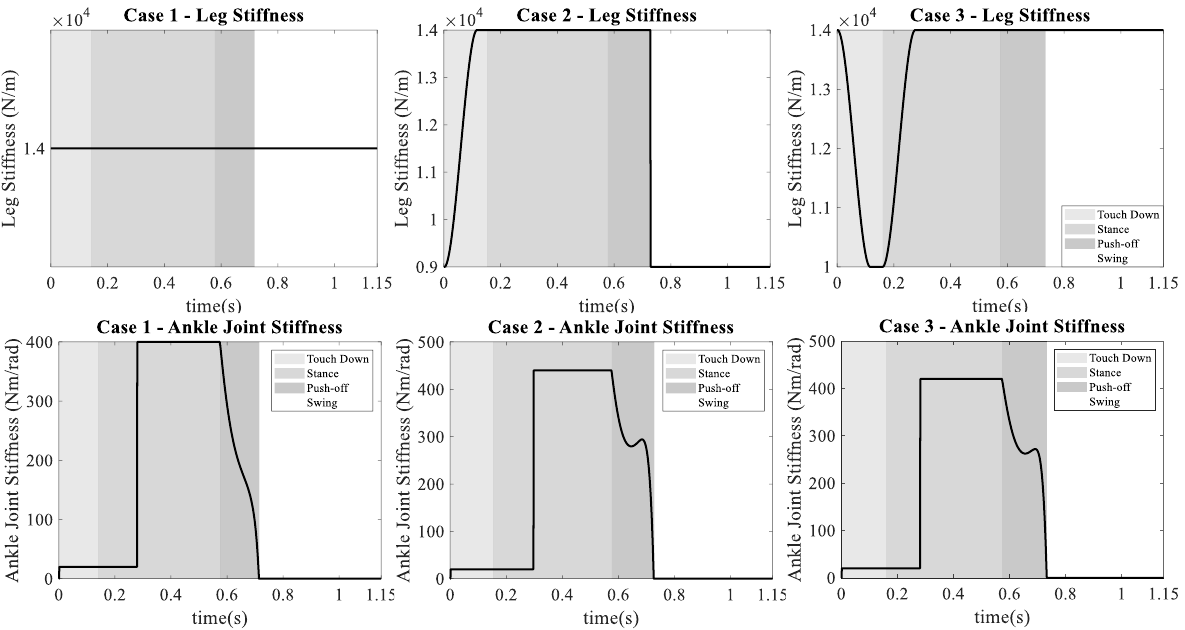}
	\caption{Stiffness profiles of compliant leg and ankle joint for each case at their distinct modes.\label{fig_7_stifProf}}
\end{figure*}

The stiffness profiles for the three cases are illustrated in Fig. \ref{fig_7_stifProf}, and the corresponding numerical values are listed in Table \ref{tab:1-ModParam}. For ankle joint stiffness, we adopt the profile pattern reported by Shamaei et al. \cite{Shamaei_ankleStiffness}, where ankle quasi-stiffness is low in early stance (\textasciitilde 200 Nm/rad) and gradually increases through mid-stance until push-off. At push-off, ankle stiffness is indirectly determined by the ankle push-off strategy described in Section 2.2. This same ankle stiffness profile is applied consistently across all three cases. Fig. \ref{fig_7_stifProf} illustrates the ankle joint stiffness profile over a full stride, where early- and mid-stance values are taken from the literature as constant quasi-stiffness within each sub-phase. %, while the push-off stiffness is obtained from successful simulation runs at the target gait speed of \textasciitilde1.2 m/s.

%%%%%%%%%%%%%%%%%  begin two column figure  %%%%%%%%%%%%%%%%%%%%%%%%%%%%%%%%%%%%

To simulate the model under these three cases, the system was implemented in MATLAB (R2024b, MathWorks Inc., Natick, MA, USA). As described in the previous section, the model incorporates four discrete modes for walking. Transitions between modes follow the event-triggered logic discussed in Section 2.3 and illustrated in Fig.~\ref{fig_8_disModeTrans}. Within each mode, the governing differential equations were solved using MATLAB’s ode45 solver, an explicit Runge–Kutta method suited for nonlinear systems. %The simulations results for all three cases are presented in Fig.~\ref{fig_9_mainResults}. %The simulation was initialized with the proposed model standing vertically, both feet in ground contact through heel and toe, and legs at their rest length. The corresponding initial condition was set as $[x_m,\ y_m,\ \dot{x}_m,\ \dot{y}_m,\ x_{hl},\ y_{hl},\ x_{hr},\ y_{hr}] = [0,\ 1.0,\ 0, \ 0, \ 0, \ 0, \ 0,\ 0,]$. Each simulation was executed until the system reached the target average forward velocity, $\dot{X}_d$, of \textasciitilde1.2 m/s and converged to a stable cyclic gait. The three stiffness cases were evaluated after the model first achieved a moderate stable gait speed (\textasciitilde 0.8 m/s), during which Raibert’s foot-placement controller gains were manually tuned to stabilize walking at \textasciitilde1.2 m/s. The simulations results for all three cases are presented in Fig. \ref{fig_9_mainResults}.

\section{Results}
%%%%%%%%%%%%%%%%%  begin two column figure  %%%%%%%%%%%%%%%%%%%%%%%%%%%%%%%%%%%%
\begin{figure*}
	\centering\includegraphics[width=0.99\linewidth]{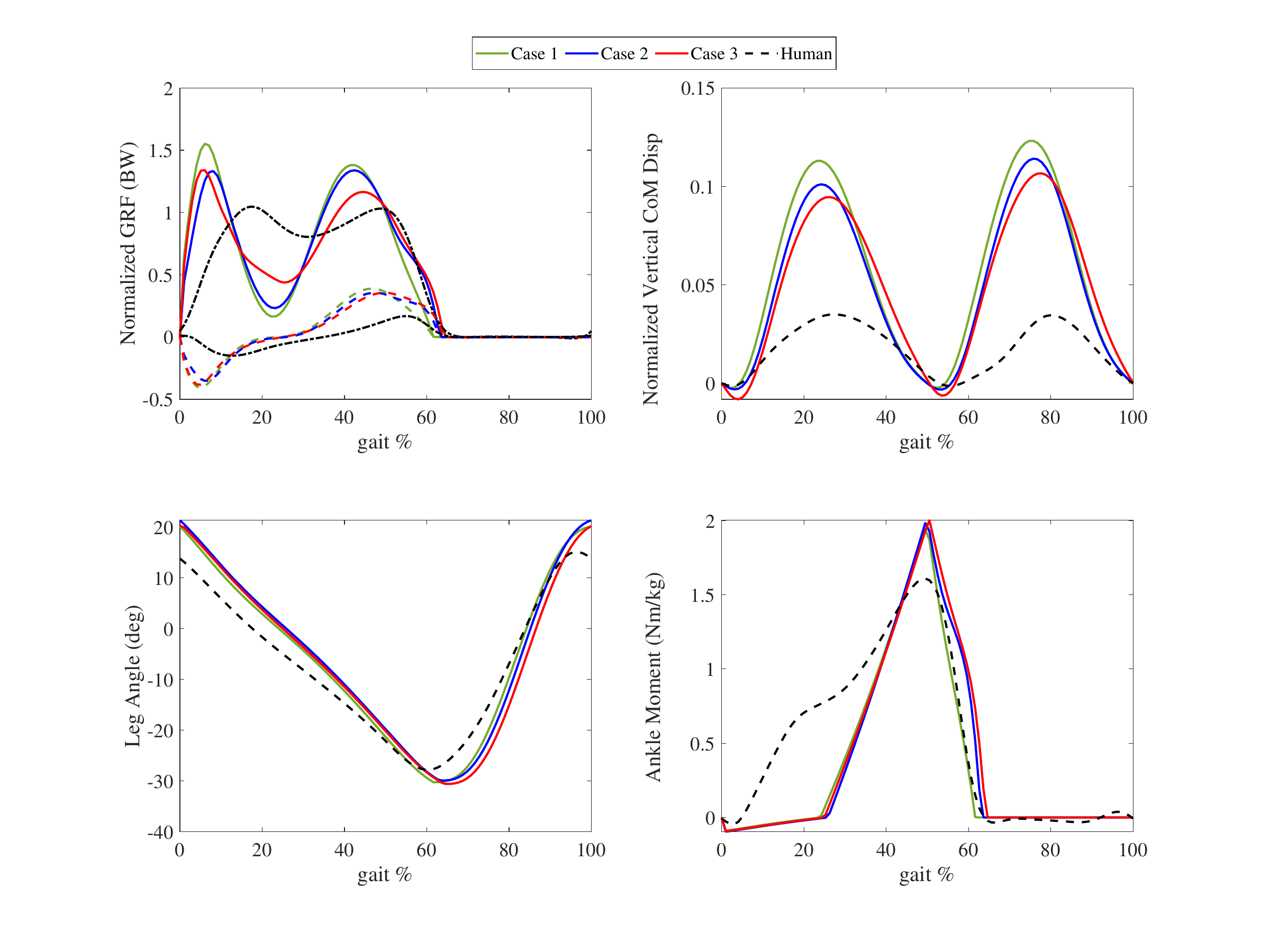}
	\caption{Walking gait characteristics at about 1.2 m/s across one cycle: simulation results for the three model cases compared with human data processed from the publicly available dataset \cite{Gregg_datasetPaper}.\label{fig_9_mainResults}}
\end{figure*}

\subsection{Proposed Model vs Human}
For comparison among the three cases of the proposed model and human data, four distinct walking characteristics were considered over one gait cycle: horizontal and vertical GRF ($GRF_x$, $GRF_y$), vertical displacements of the CoM ($\Delta Y_{CoM}$), leg angle ($\theta_i$), and ankle joint moment ($\tau_{ai}$), as shown in Fig. \ref{fig_9_mainResults}. %Note that the objective of this paper is to propose a new simplified model that demonstrates robustness and captures human-like characteristic patterns in walking, and thus the parameter optimization was not employed to apply tuned stiffness and damping values for exact quantitative agreement.

Across all cases, it is noticeable that $GRF_x$ transitions from negative to positive values during the gait cycle, and $GRF_y$ exhibits the characteristic M shaped profile, which are consistent with human walking data. Similarly, the $\Delta Y_{CoM}$ oscillates about its mean height with maximum displacement at mid-stance. While these characteristics qualitatively match human data, quantitative discrepancies remain. As noted in \cite{Wensing_variStiffness}, achieving closer agreement of these characteristics requires continuously varying leg stiffness throughout the gait cycle, rather than assigning constant stiffness values to each distinct phases. Even though no parameter optimization was performed and the parameters were selected using normative values in humans, our model reproduces human-like patterns in GRF, CoM vertical displacement, and ankle moment while maintaining stable cyclic walking. 

The ankle moment, apart from the first 20\% of the gait cycle, is similar to the human profile. The reason for small ankle moment right after touchdown was the small ankle stiffness adopted for this phase (Fig. \ref{fig_7_stifProf}). We found that larger touchdown stiffness for the ankle affects the stability of walking.

Another important characteristic to consider is gait timing. Based on the human data from the utilized datasets for walking at speed of 1.2 m/s, the touchdown subphase of the stance foot lasts for approximately the first 10–15\% of the gait cycle, mid-stance lasts for about 35–40\%, push-off begins around 50\%, and the swing phase starts between 65\% and 70\% of the gait cycle. When examining the gait timing of all three cases of the proposed model, it is confirmed that the model follows approximately the same timing for walking at 1.2 m/s, as indicated by the shaded regions in gray and white in Fig. \ref{fig_7_stifProf}. Taken together with these analyzed characteristics, the results show that the proposed model is capable of reproducing human-like walking features, including anthropomorphic characterization of hip and ankle joints as well as swing-foot behavior. Furthermore, consistent with previous literature \cite{Wensing_variStiffness, Shamaei_ankleStiffness}, the results indicate that time-varying stiffness profiles produce gait characteristics more closely aligned with human walking than those obtained with constant stiffness. On the other hand, as it was also observed in \cite{Geyer_SLIPwalkRun} and \cite{Rezazadeh_Hurst_2020}, the peak GRFs and CoM displacements can be significantly larger than those of humans.

\subsection{Comparison with the Point-Foot Model}
%We also implemented and tested the D-SLIP model to enable a direct comparison with the proposed DVSLIP-FF model. In this paper, we present simulation results for both models, obtained using the same parameter set corresponding to Case 2 : a varying leg stiffness profile characterized by low stiffness at touchdown that gradually increases to a higher value at mid-stance.

\begin{figure}
	\centering\includegraphics[width=.99\linewidth]{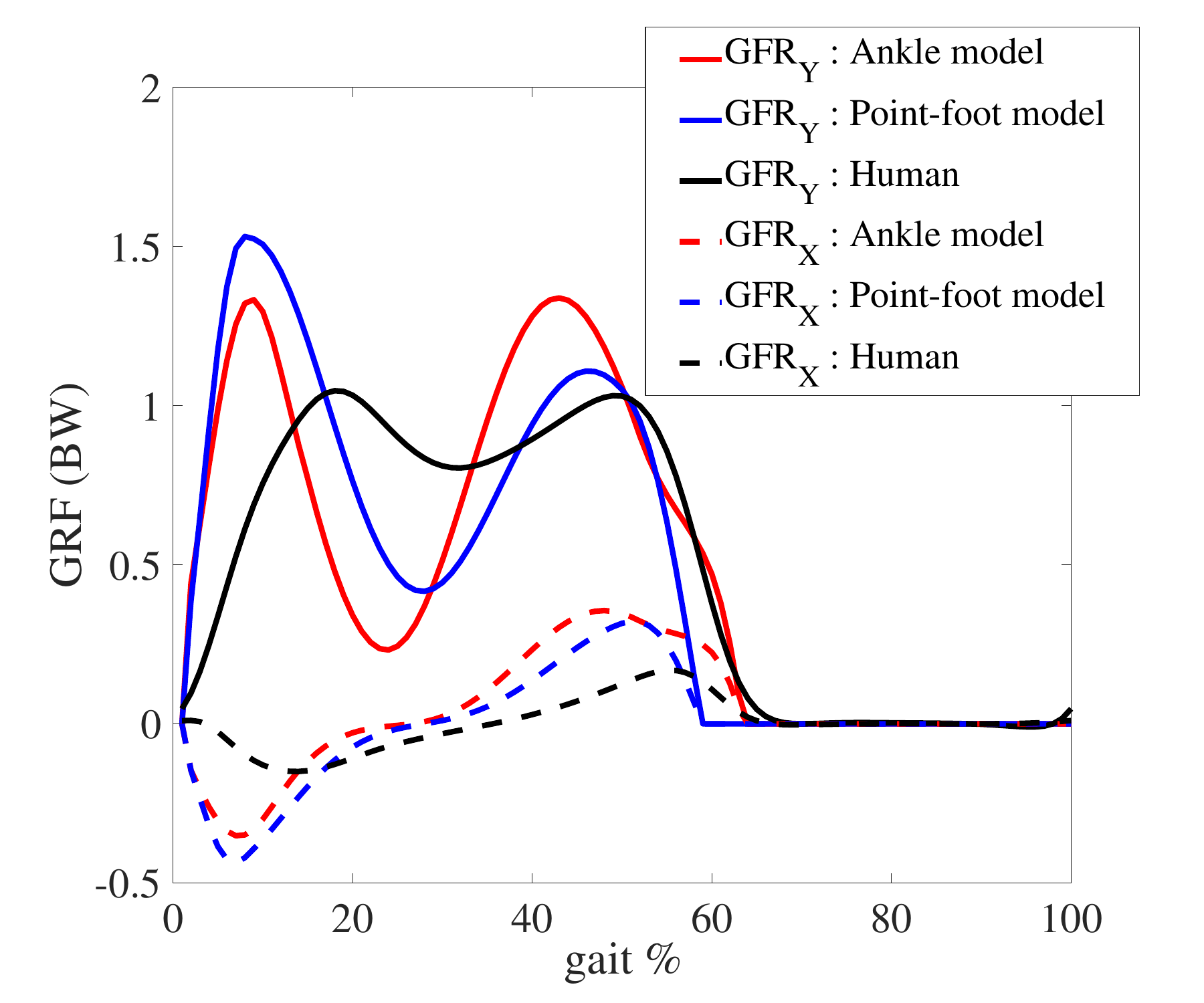}
	\caption{Comparison of simulation results for the model with point-feet and the model with ankles, with human walking data.\label{fig_10_modelComp}}
\end{figure}

Figure \ref{fig_10_modelComp}, presents the results for both the point-foot and ankle-controlled cases. Both models reproduce qualitatively similar GRFs. However, in the new model, $GRF_Y$ exhibits two nearly equal peaks, producing a more symmetric profile as observed in humans. In contrast, the point-foot model displays a noticeably higher first peak compared to the second. Further, during the push-off phase, the new model aligns more closely with the human trajectories than the point-foot. Moreover, the new model demonstrates better performance in reproducing human-like gait timings, specifically through the prolonged stance phase.

%%%%%%% SINGLE COLUMN FIGURE

\subsection{Enhanced Walking Stability by the Proposed Ankle Push-off}
As discussed in Section 2.2, we utilize a velocity-based foot placement scheme to control the desired touchdown position, $\Delta X_{hi}$, of the swing foot relative to the CoM. As proven in \cite{Rezazadeh_Hurst_2020}, in the point-foot model, this foot placement scheme is critical for the stabilization of the gait cycle, even for small perturbations. However, observations in humans indicate that for small perturbations, foot placement is not regulated and instead, ankle strategies are employed to maintain stability \cite{vlutters2016center}. In this subsection, we show that the proposed ankle push-off scheme provides a similar outcome; i.e., unlike the point-foot model, it does not need foot placement adjustments to maintain the local gait stability against small perturbations.

%%%%%%%%%%%%%%%%%  begin two column figure  %%%%%%%%%%%%%%%%%%%%%%%%%%%%%%%%%%%%
\begin{figure*}
	\centering\includegraphics[width=0.9\linewidth, trim={0mm 25mm 0mm 30mm}]{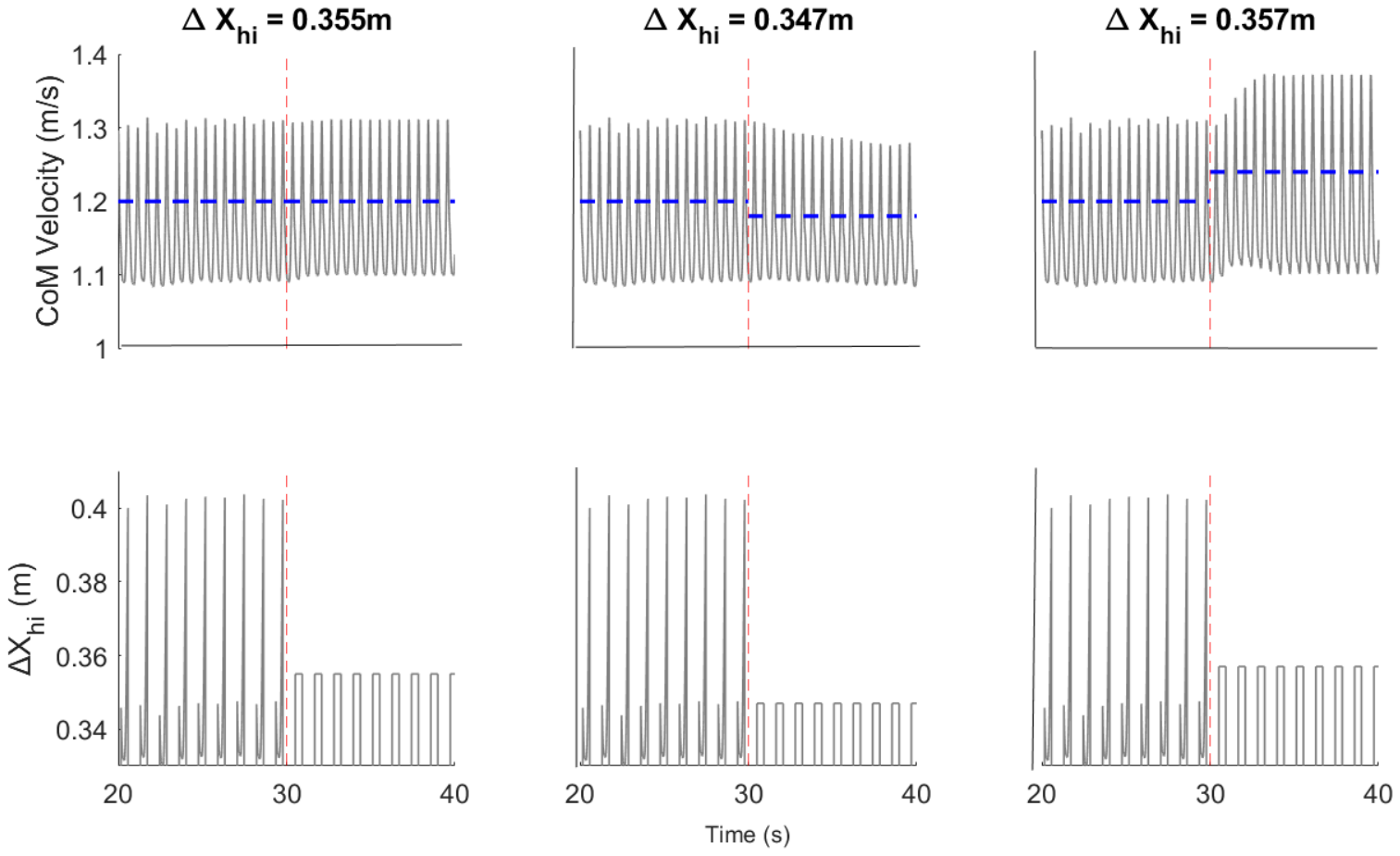}
	\caption{Stability of the model without foot placement control, which is turned off at $t=30"$~s (vertical red line). When the foot placement is fixed to a slightly different value of that of the original gait cycle (middle and right columns), the system converges to different gait cycles with slightly different average velocities (dashed blue lines).   \label{fig_11_swingCon}}
\end{figure*}

To demonstrate this, we simulated the system starting with the controlled foot placement and after a number of steps (once convergence was ensured), we switched to unregulated fixed foot placement. The system did not fail by stumbling or falling, but continued stable walking on a new periodic orbit with a slightly different average gait speed. As shown in Fig.~\ref{fig_11_swingCon}, this was repeated three times with switching from foot placement regulation to fixed foot placement with three different values, and each time the system converged to a different stable cycle. These results confirm that the proposed push-off paradigm can sustain local stability of the cycle in the absence of foot placement control. This capability is not observed in other SLIP-family models, which typically fail to maintain stable walking when the swing-foot placement control is disabled \cite{Vejdani_Geyer_Hurst2025}.

%%%%%%%%%%%%%%%%%%%%%%%%%%%%%%%%%%%%%%%%%%%%%%%%%%%%%%%%%%%%%%%%%%%%%%
\section{Discussions}
We examined the addition of a foot-ankle system to the forced-oscillation-based reduced-order model, which was previously analyzed only with point feet. The significance of this reduced-order model lies not only in its ability to capture qualitative features of anthropomorphic walking but also in its high stability, which can facilitate implementation on robotic systems and potentially provide insights into stability mechanisms of human walking. We performed various simulations to test both of these sets of characteristics. Since our goal was to assess the overall behavior of the system with normative values from human locomotion, we did not use trajectory optimization methods. Nonetheless, most of the outcome characteristics of the model with the ankle–foot addition were qualitatively similar to their human counterparts, while also being quantitatively slightly closer than those of the point-foot model.  This was also demonstrated in the gait timings of the two models compared to those of humans, where the new model produced a closer matching. This was mainly due to the added push-off which prolongs the stance phase. Most importantly, our focus extended beyond reproducing human-like trajectories to also include the stability characteristics. In particular, we showed that the proposed ankle characteristics can replicate the stabilization strategies observed in humans, where ankle control is used to reject small disturbances rather than relying on foot placement regulation for such cases. Note that the velocity-based foot placement control, originally designed for the point-foot forced-oscillation-based model, was still effective in controlling the new model under large initial condition errors. As such, equipping the forced-oscillation-based reduced-order model with an ankle–foot module not only provides the incorporation of these important features, but also helps in achieving a more accurate replication of the human walking stability and dynamics characteristics. 

Although we intentionally avoided trajectory optimization methods in our work to focus on human-inspired parameters, we did adjust the stiffness according to both human data and the optimization-based findings in the previous works. The variable stiffness models succeeded in slightly reducing the peaks in GRFs, although not enough to bring the forces close to the human levels. Among three cases, the results in Fig. \ref{fig_9_mainResults} show that Case 2 produces a more symmetric M-shaped $GRF_y$. This aligns with prior studies \cite{Wensing_variStiffness, Zhao_varyStiff}, confirming that variable stiffness models (Cases 2 and 3) better approximate human walking than the constant stiffness model (Case 1). Notably, the larger variations of the GRFs (especially in the horizontal direction) and the CoM vertical displacements were also observed in the experiments with ATRIAS, which was controlled through the point-foot version of the forced-oscillation-based walking control \cite{Rezazadeh_Hurst_2020}. Likewise, Geyer et al. observed that in bipedal SLIP, the decrease of the touchdown angle would result in a similar increase in the peak GRFs compared to that of humans \cite{Geyer_SLIPwalkRun}. However, note that changing the touchdown angle with the same forward velocity is equivalent to changing the gait period. This implies that matching the model’s gait period, touchdown angle, and GRFs to those of humans requires some form of stance-phase control and/or variable stiffness, as observed in \cite{Wensing_variStiffness} for the SLIP model. Nonetheless, the essential advantages of SLIP compared to inverted pendulum model of walking (particularly, force and CoM displacement patterns \cite{Geyer_SLIPwalkRun}) were also preserved in both point-foot and ankle-foot forced-oscillation-based models. However, the human-inspired stability features, both in foot placement and in ankle control, provide a significant progress compared to the basic bipedal SLIP model. These features can not only enhance the understanding of human locomotion but also facilitate the application of the reduced-order model to bipedal robots.

\section{Conclusions}
We investigated the addition of an ankle and foot system to the previously proposed point-foot forced-oscillation-based model of bipedal walking. We showed that human-inspired control schemes for ankle dorsiflexion and its push-off plantarflexion would lead to qualitatively similar human-like outcomes in GRFs and CoM displacement as bipedal SLIP model. In addition and more importantly, we demonstrated that this ankle control scheme proprioceptively rejected small disturbances without adjusting foot placement, which is consistent with observations in humans. This can potentially help in improved energy efficiency through eliminating the control effort for the foot placement regulations. 

The future works will focus on two main topics. First, we will investigate how the ankle control can be activated immediately after touchdown, with the goal of resolving the slight dependency between the ankle moment in the model and in the humans. Second, we will investigate the continuously variable stiffness profiles both for the leg and for the ankle, in order to achieve more human-like outcomes, especially in the peak GRFs. This can be achieved for example by combining trajectory optimization techniques with human quasi-stiffness profiles.

%For future work, we plan to extend this study by incorporating time-varying stiffness profiles for both compliant leg and the ankle joint in order to achieve quantitative agreement with human walking characteristics. Furthermore, we aim to integrate trajectory optimization to the proposed model, to enable more efficient and agile locomotion with improved motion accuracy (closely following desired trajectories), and enhanced stability against external perturbations.

\bibliographystyle{IEEEtran}
\bibliography{bibliography}

\end{document}